\documentclass[letterpaper, 10 pt, journal, twoside]{IEEEtran}  %
\pdfoutput=1
\usepackage{amsmath} %
\usepackage{amssymb}  %
\usepackage{graphicx} %
\usepackage[acronym]{glossaries}
\usepackage{booktabs}

\usepackage{enumitem}

\usepackage[breaklinks,colorlinks]{hyperref}
\usepackage{siunitx}
\usepackage[ruled,vlined]{algorithm2e}
\usepackage{algorithmic}
\usepackage{microtype}
\usepackage{flushend}
\usepackage[hang,flushmargin]{footmisc}

\usepackage{url}
\usepackage[dvipsnames]{xcolor}
\usepackage[export]{adjustbox}

\usepackage{pifont}%
\usepackage{siunitx}
\usepackage{cuted}
\usepackage{threeparttable}
\usepackage{multirow}
\usepackage{cite}
\usepackage{siunitx} %

\usepackage{tikz}
\usepackage{pgfplots}
\pgfplotsset{compat=1.18}

\usepackage[resetlabels]{multibib}
\newcites{S}{References}
\usepackage{makecell}
\usepackage{array}    %
\usepackage{moresize} %
\usepackage{tabularx}
\usepackage{comment}
\newcolumntype{Y}{>{\centering\arraybackslash}X}
\newcolumntype{Z}{>{\raggedleft\arraybackslash}X}

\definecolor{dark-green}{RGB}{12,80,12}

\newcommand{\secref}[1]{Sec.~\ref{#1}}
\renewcommand{\eqref}[1]{Eq.~(\ref{#1})}
\newcommand{\figref}[1]{Fig.~\ref{#1}}
\newcommand{\tabref}[1]{Tab.~\ref{#1}}

\DeclareMathOperator*{\argmax}{arg\,max}
\DeclareMathOperator*{\mdp}{MDP}

\newcommand{\mmvec}[3]{(\SI{#1}{\mm}, \SI{#2}{\mm}, \SI{#3}{\mm})}

\newcommand{\para}[1]{\parskip=5pt\noindent\textit{#1}}
\newcommand{\ours}{MoMa-CoDesign}
\newcommand{\ourslong}{Task-Driven Co-Design of Mobile Manipulators}
\newcommand{\website}{\url{https://moma-codesign.cs.uni-freiburg.de}}

\newcommand{\myworries}[1]{{#1}}

\title{\ourslong{}
}

\author{Raphael Schneider$^*$, Daniel Honerkamp$^*$, Tim Welschehold, and Abhinav Valada %
\thanks{$^{*}$ Equal contribution. All authors are with the Department of Computer Science, University of Freiburg, Germany.}%
\thanks{This work was partially funded by the German Research Foundation (DFG): 417962828, an academic grant from NVIDIA, and the BrainLinks-BrainTools center of the University of Freiburg.}%
\thanks{© 2025 IEEE.  Personal use of this material is permitted.  Permission from IEEE must be obtained for all other uses, in any current or future media, including reprinting/republishing this material for advertising or promotional purposes, creating new collective works, for resale or redistribution to servers or lists, or reuse of any copyrighted component of this work in other works.}%
}

\begin{document}
\newacronym{ac:bohb}{BOHB}{Bayesian Optimization and HyperBand }
\newacronym{ac:rl}{RL}{Reinforcement Learning }
\newacronym{ac:sac}{SAC}{Soft Actor-Critic }
\newacronym{ac:er}{ER}{Evolutionary Robotics }
\newacronym{ac:drl}{DRL}{Deep Reinforcement Learning }
\newacronym{ac:bo}{BO}{Bayesian Optimization  }
\maketitle
\begin{abstract}
Recent interest in mobile manipulation has resulted in a wide range of new robot designs. A large family of these designs focuses on modular platforms that combine existing mobile bases with static manipulator arms. They combine these modules by mounting the arm in a tabletop configuration. However, the operating workspaces and heights for common mobile manipulation tasks, such as opening articulated objects, significantly differ from tabletop manipulation tasks. As a result, these standard arm mounting configurations can result in kinematics with restricted joint ranges and motions.
To address these problems, we present the first Concurrent Design approach for mobile manipulators to optimize key arm-mounting parameters. Our approach directly targets task performance across representative household tasks %
by training a powerful multitask-capable reinforcement learning policy in an inner loop while optimizing over a distribution of design configurations guided by \gls{ac:bohb} in an outer loop.
This results in novel designs that significantly improve performance across both seen and unseen test tasks, and outperform designs generated by heuristic-based performance indices that are cheaper to evaluate but only weakly correlated with the motions of interest. %
We evaluate the physical feasibility of the resulting designs and show that they are practical and remain modular, affordable, and compatible with existing commercial components. %
We open-source the approach and generated designs to facilitate further improvements of these platforms.

\end{abstract}
\glsresetall

\section{Introduction}
\begin{figure}[t]
	\centering
	\resizebox{\linewidth}{!}{%
 \includegraphics[width=\linewidth,trim={0cm 0cm 0cm 1cm},clip,angle =0,valign=c]{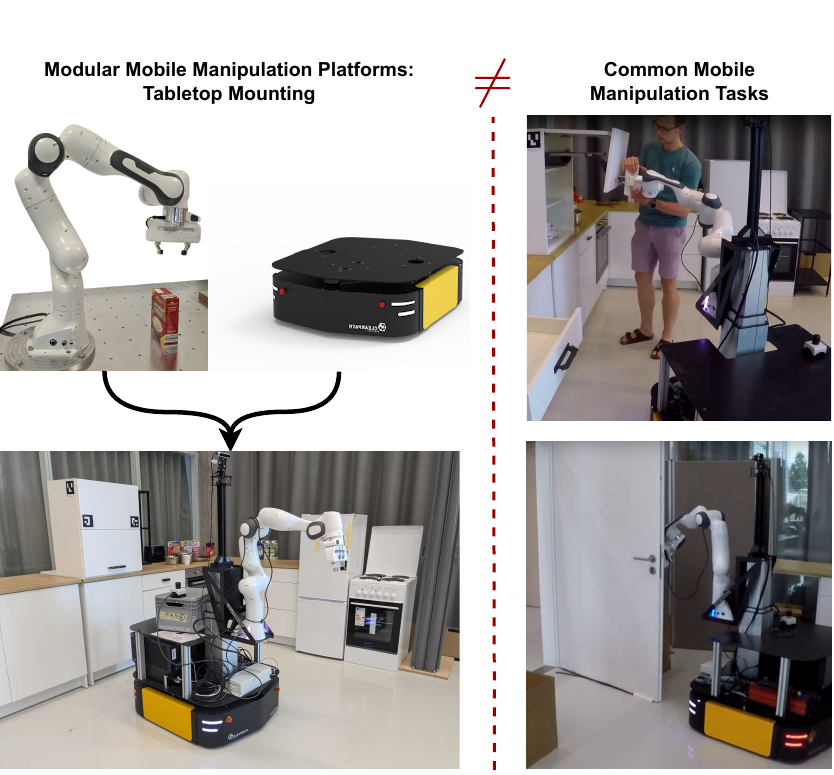}}
     \caption{Modular platforms have become a popular, cost-efficient design for mobile manipulators. However, they typically mount the manipulator arms identically to tabletop settings. We investigate this design choice and find that it is suboptimal for common mobile manipulation tasks that require different sets of motions. We then present a concurrent design approach to optimize the mounting parameters directly based on task success.\looseness=-1} 
  	\label{fig:teaser}
\vspace{-0.5cm}
\end{figure}

\IEEEPARstart{I}{n} In the last few years, we have seen an increasing adoption of mobile manipulator robots, leading to a wide range of new robot designs. Most commonly, mobile manipulators are constructed as modular systems based on existing commercially available mobile bases and manipulator arms, as shown in \figref{fig:collage}. This is a simple and cost-efficient strategy. However, we find that these platforms overwhelmingly follow the simplest possible mounting strategy, placing the manipulator arm on the base in the same configuration as for tabletop manipulation.
In this work, we evaluate this design choice across a wide range of mobile manipulation tasks and find it to often lead to suboptimal kinematics, with restrictive joint limits that make it difficult to perform mobile manipulation tasks that often require different motions and workspaces than tabletop task, such as sideways grasping of drawers and handles at various heights, as illustrated in \figref{fig:teaser}.

\setlength{\tabcolsep}{1pt}
\setlength{\tabcolsep}{6pt}

Concurrent Design (co-design)~\cite{doncieux2015evolutionary, bongard2013evolutionary} simultaneously refines the robot’s morphology and control policy, yielding more holistic and high-performing solutions. %
These methods have been applied to systems ranging from locomotion~\cite{bravo2020one, dinev2022versatile} to manipulator arms~\cite{kulz2024optimizing, hwang2017design}. %
In this work, we identify the mounting parameters of mobile manipulators as a crucial design element and propose the first approach for the co-design of modular mobile manipulators. While our approach is not limited to these parameters, by focusing on arm mounting, the generated designs retain modularity and cost-efficiency, making them easily realizable in the real world.

Having identified the relevant parameters, we propose to directly optimize task performance by using the success rate on representative mobile manipulation tasks as a scoring function.
This is made possible by the use of a powerful reinforcement learning policy, N$^2$M$^2$~\cite{honerkamp2023n}, that trains a mobile manipulator to enable arbitrary end-effector motions. Importantly, this approach was shown to generate effective whole-body motions across a wide range of robot kinematics and tasks~\cite{honerkamp2024zero, honerkamp2021learning,honerkamp2024language,schmalstieg2023learning} and can be trained without an expensive physics simulator. In an outer loop, we maintain a distribution over possible designs and use \gls{ac:bohb}~\cite{falkner2018bohb} to efficiently sample promising regions of the high-dimensional design space.
In contrast to existing work that largely focuses on heuristic scoring functions or optimizes for very specific tasks~\cite{hwang2017design, dinev2022versatile, xu2021end}, we hypothesize that this direct optimization for task performance with a general policy will result in designs and policies that generalize to the diverse tasks and environments that mobile manipulators are confronted with in human-centered environments. 
We evaluate the performance of the resulting designs on unseen test tasks and compare our approach to heuristic-based performance indices that allow the evaluation of a much larger number of designs but are not directly task-aligned.
We find that the \myworries{designs generated by all proposed methods yield} significantly better performance than the default tabletop-style arm mounting. Moreover, the proposed direct task-based optimization consistently yields the best performance across different base drives and arms.
We find that the designs generated (or optimized) by all the
proposed methods yield significant better performance than the default tabletop-style mounting.

To summarize, this work makes the following contributions:
\begin{itemize}
    \item We identify the arm mounting parameters as a crucial design element in common modular mobile manipulators. 
    \item We develop the first co-design method for mobile manipulators and demonstrate significant design improvements that generalize across a wide range of tasks.
    \item We develop a directly task-driven objective for mobile manipulation and show that its direct optimization outperforms heuristic-based design metrics for manipulators.
    \item We demonstrate the methodology's versatility across different robot models and show that the approach can improve the designs of a wide range of modular mobile manipulator platforms.
    \item We make the code publicly available at \website{}.
\end{itemize}

\section{Related Work}

\begin{figure}[t]
    \centering
	\resizebox{\linewidth}{!}{%
	\begin{tabular}{cccccc}
        \includegraphics[width=\linewidth]{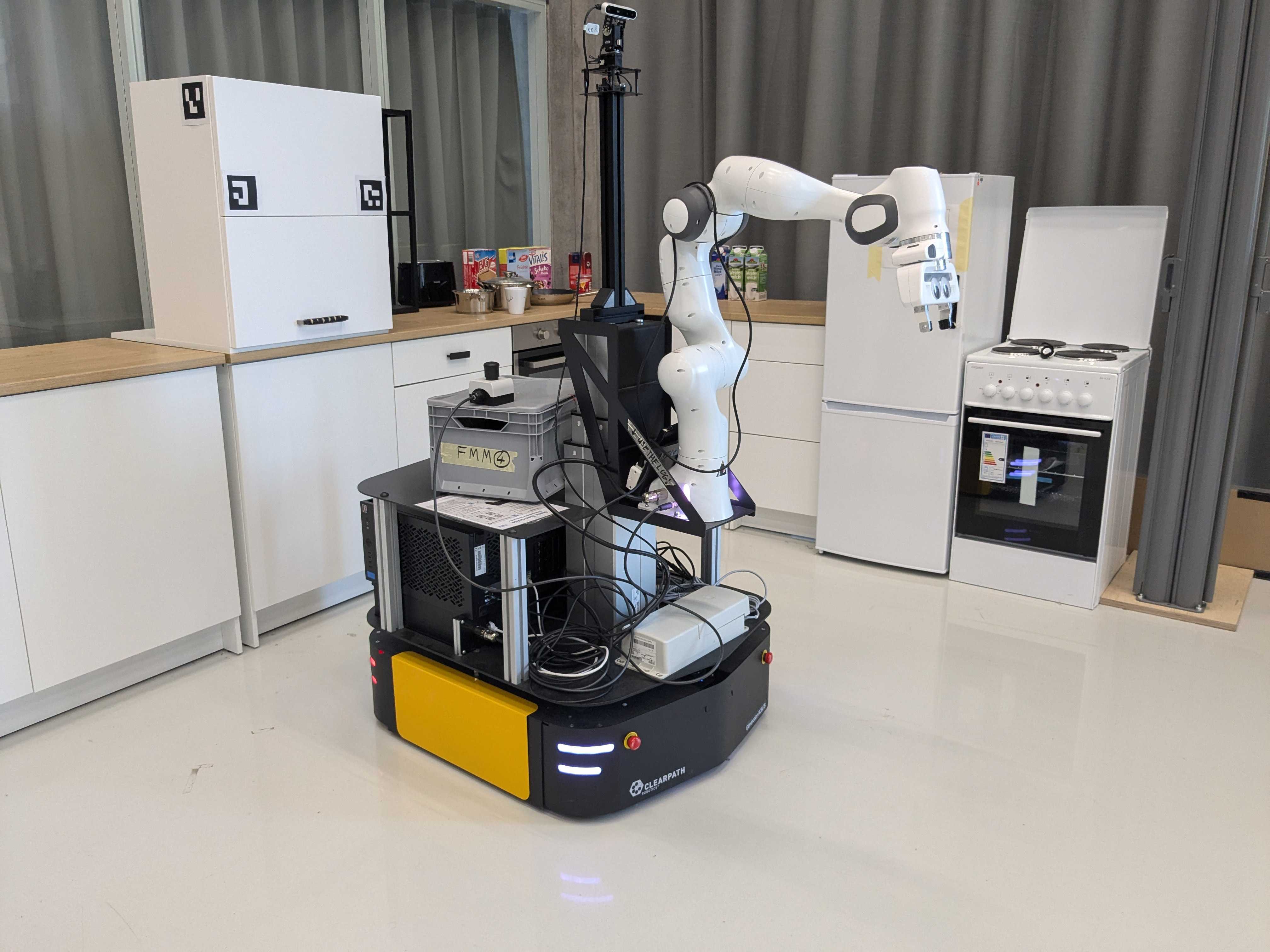} &
        \includegraphics[width=\linewidth]{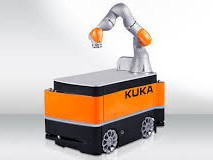} &
        \includegraphics[width=\linewidth]{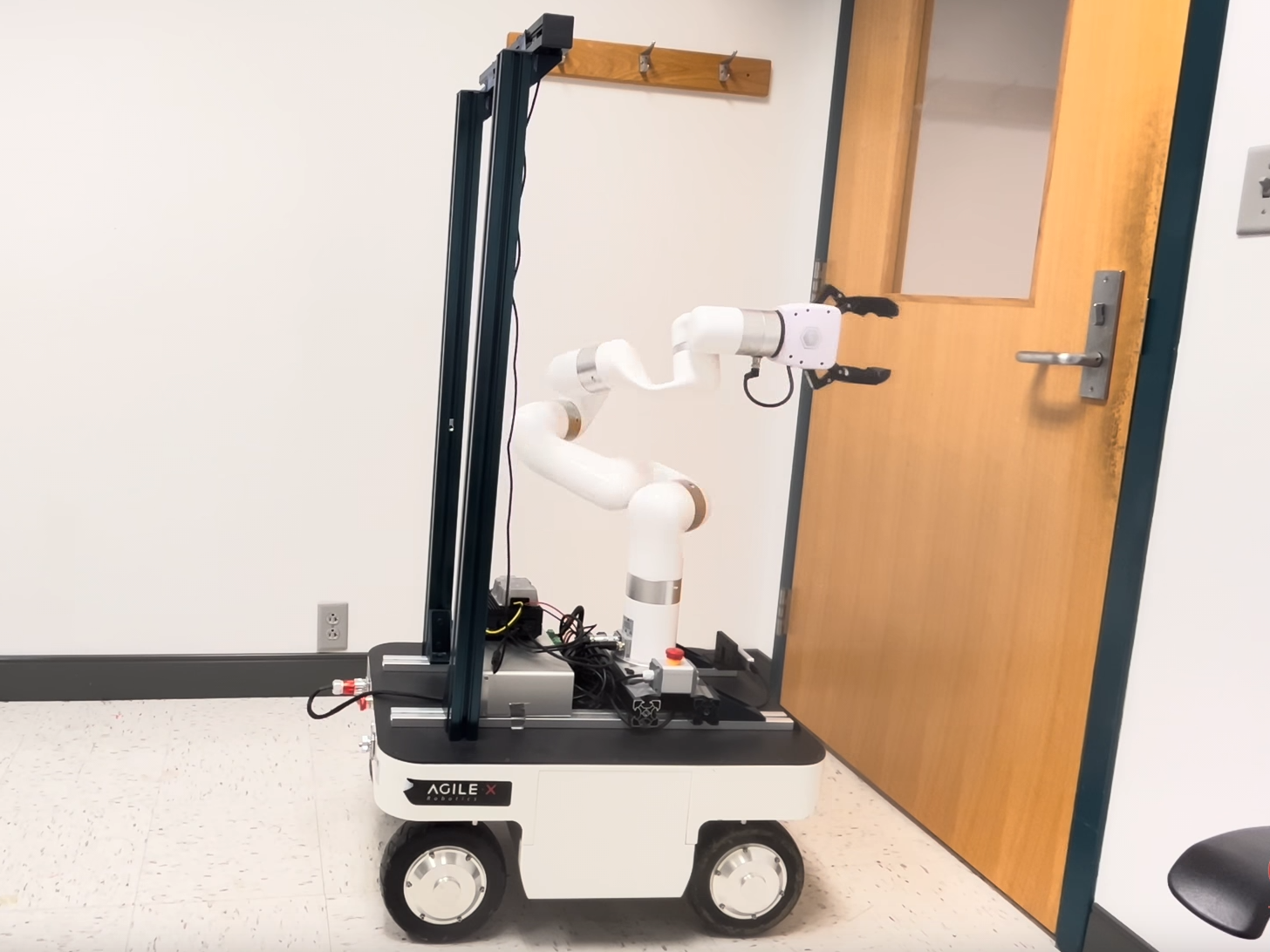} \\
        \includegraphics[width=\linewidth]{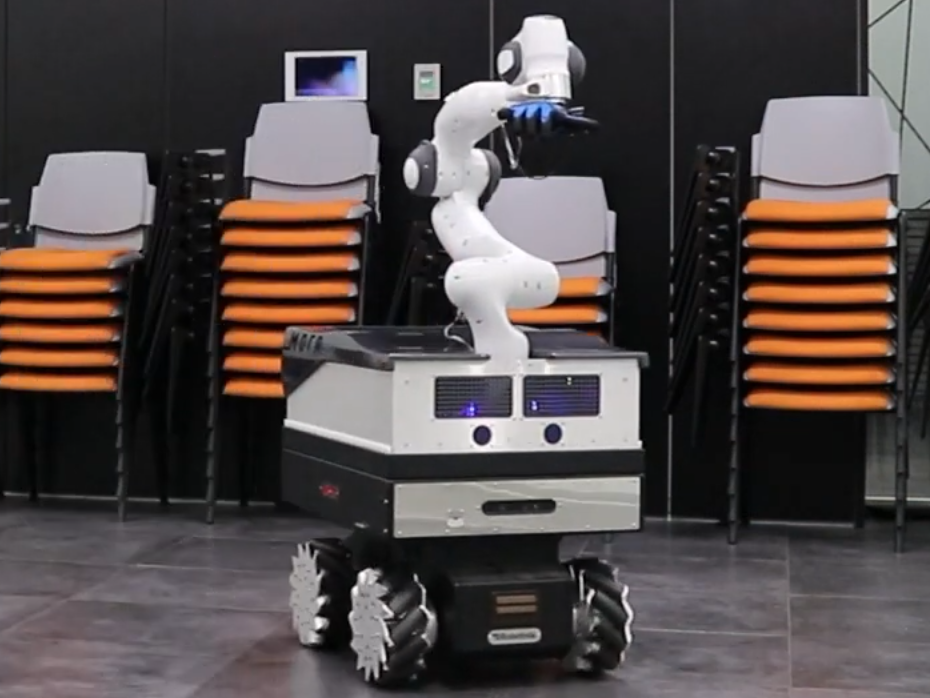} &
        \includegraphics[width=\linewidth]{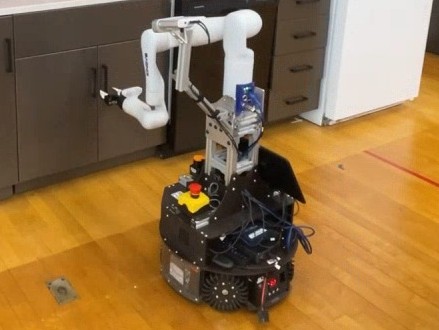} &
        \includegraphics[width=\linewidth]{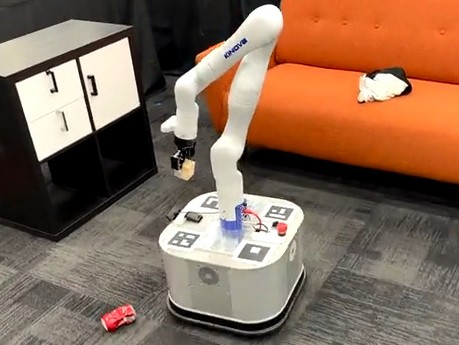}\\
	    \end{tabular}
        }
    \caption{A wide range of mobile manipulator designs have been developed. Most commonly, they follow a modular design and mount existing manipulator arms in a tabletop configuration on existing mobile bases. References from top left to bottom right~\cite{honerkamp2024zero,kmr_iiwa,xiong2024adaptive,raei2024multipurpose,hsu2024kinscene,wu2023tidybot}%
    .}
    \label{fig:collage}
   \vspace{-0.3cm}
\end{figure}

{\parskip=2pt
\noindent\textit{Mobile Manipulator Designs}: 
Mobile manipulators can take a wide range of forms. Commercial platforms often follow human-inspired designs with shoulder-mounted arms, such as the PR2 or TIAGo. However, these designs created from scratch come at high costs and details on their design decisions are not publicly available. While Wu~\textit{et~al.}~\cite{wutidybot} open-source the base design of a mobile manipulator, they do not study the arm design. As a result, modular mobile manipulator designs consisting of widely available manipulator arms on existing mobile bases have gained large adoption, as shown in \figref{fig:collage}. However, these robots overwhelmingly mount manipulator arms in a tabletop configuration, not taking into account the differences between mobile manipulation and static manipulation tasks. While some different arm mounts exist, their design is driven to enable human teleoperation~\myworries{\cite{schwarz2023robust, hauser2024analysis}} or directly given by the arm, such as the shoulder-mounting inspired ABB YuMi arms~\cite{blomqvist2020go} without further considerations.}
{\parskip=2pt
\noindent\textit{Concurrent Design}
jointly optimizes both the physical design and control strategies to enhance the robot's performance.
A wide range of methods have been proposed to tackle this problem in various settings. Derivative-based methods have been used to optimize design and motion policies in a single nonlinear program~\cite{spielberg2017functional}. Gradients obtained via sensitivity analysis have enabled the optimization along the manifold of the optimal motions~\cite{ha2017joint} and the use of hard constraints~\cite{dinev2022versatile}.
In parallel, sampling-based methods such as Bayesian Optimization~\cite{liao2019data} and evolutionary algorithms~\cite{digumarti2014concurrent, gupta2021embodied, kulz2024optimizing} have been developed to improve the fitness of functions over time. Reinforcement learning has been used to learn control policies in the inner loop~\cite{chen2020hardware}, design parameters in the outer loop~\cite{albright2022selecting}, or for more efficient overall co-design by learning value functions over design parameters~\cite{luck2020data}, a single policy across designs~\cite{schaff2019jointly, schaff2022soft} or optimizing both design and policy parameters jointly by differentiating through the design parameters~\cite{chen2020hardware, jackson2021orchid, xu2021end}. While these methods have been applied to complex problems, such as quadruped locomotion with high numbers of degrees of freedom, these methods have overwhelmingly focused on the optimization for single tasks such as forward locomotion~\cite{digumarti2014concurrent, luck2020data, jackson2021orchid, spielberg2017functional, gupta2021embodied}, jumping~\cite{dinev2022versatile}, moving an object forward~\cite{ha2017joint} or dexterous hand grasping~\cite{chen2020hardware}. Stochastic programming has incorporated optimization over an uncertain distribution of scenarios~\cite{bravo2020one, bravo2022large}. Dong~et~al.~\cite{dong2024co} combine Bayesian Optimization over task and design pairs with reinforcement learning.

Within manipulation, global performance indices such as manipulability~\cite{yoshikawa_manipulability} are commonly used to measure the fitness of robotic arms~\cite{xiao2016manipulator, molaei2022kinematic, hwang2017design, feng2020multi}. However, these indices do not directly capture task performance and can only serve as heuristics. Vaish~\textit{et~al.}~\cite{vaish2024co} use environmental constraints to efficiently optimize a 3~degree of freedom (DoF) arm for a writing task as measured by vertical force and horizontal velocity resolution\myworries{, however only investigate fixed directional constraints}. Xu~\textit{et~al.}~\cite{xu2021end} individually optimize a manipulator for tasks such as flipping a box or rotating a cube. K\"ulz~\textit{et~al.}~\cite{kulz2024optimizing} optimize complex arm designs for collision-free trajectories across a range of synthetic goal-reaching tasks. In contrast, we investigate designs of redundant mobile manipulators for tasks that require complex environment interactions such as opening articulated objects for which finding optimal policies for a single given design is already a challenging problem~\cite{pankert2020perceptive, honerkamp2023n}.
We, therefore, conceptually follow a black-box approach that uses a decoupled inner policy optimization loop with a sampling-based outer loop to maintain a distribution over designs~\cite{gupta2021embodied, schaff2019jointly, schaff2022soft, luck2020data}, and directly optimize for task success across representative tasks instead of weaker correlated heuristics.
To the best of our knowledge, this work is the first to \myworries{systematically} optimize mobile manipulator designs beyond the robot base~\cite{wutidybot}. \myworries{AVATRINA~\cite{correia2024immersive} evaluate arm and gripper mounting candidates for a mobile avatar, however they focus on a small set of candidates and do not include the base mobility in the evaluation metrics.}

\section{Design Optimization for Mobile Manipulators}%
\setlength{\tabcolsep}{1pt}
\begin{figure}[t]
	\centering
 \includegraphics[width=\linewidth,trim={0.7cm 0cm 1cm 0cm},clip,angle =0,valign=c]{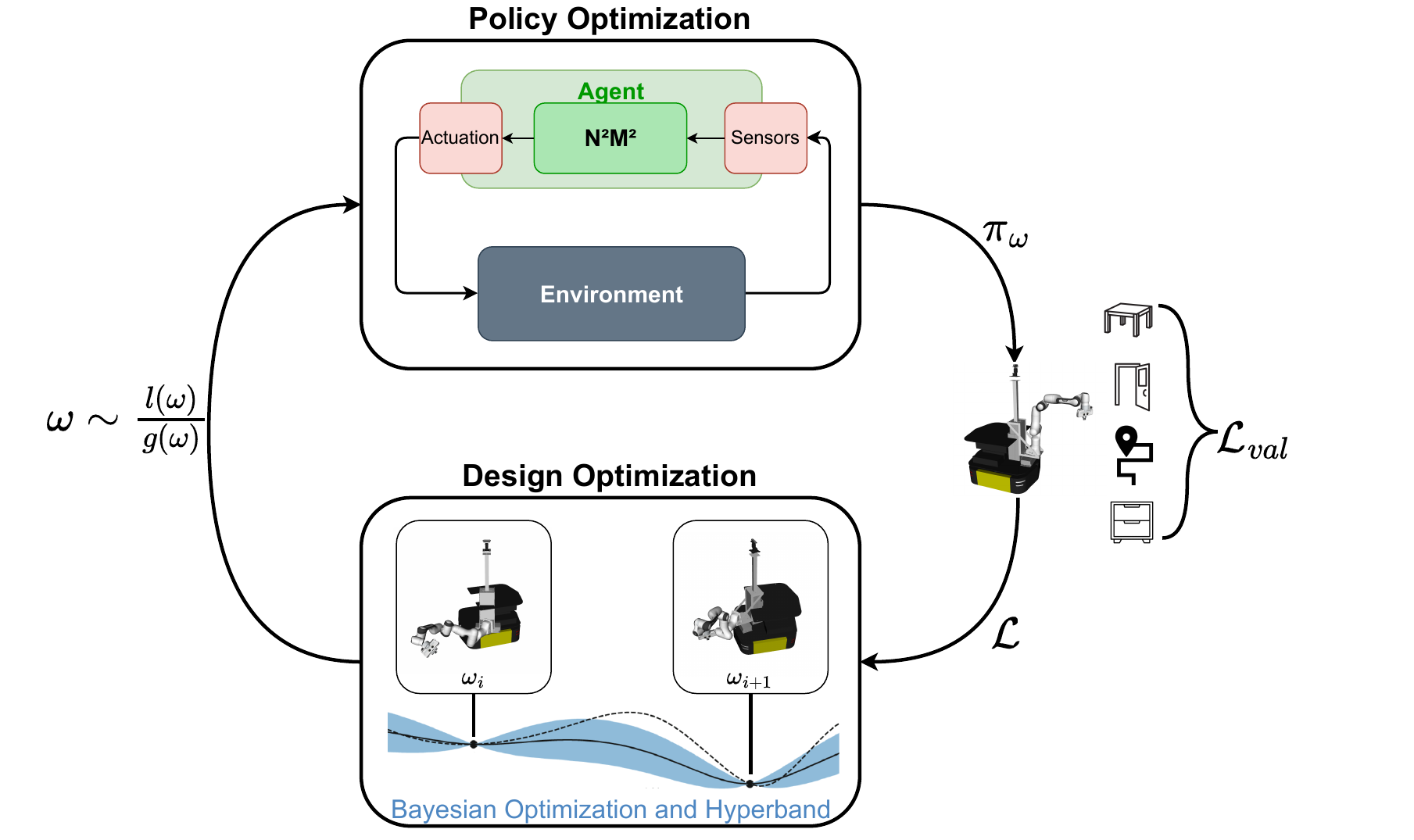}
     \caption{\ours{} maintains a distribution over design optimality based on Bayesian Optimization and Hyperband to propose new design parameters $\omega$ for evaluation. In an inner loop, we use a powerful reinforcement learning agent to learn a policy for this design. We then propose to directly use task-based success rates as a scoring function $\mathcal{L}$ to evaluate the design.\looseness=-1} 
  	\label{fig:overview}
\vspace{-0.2cm}
\end{figure}
\setlength{\tabcolsep}{6pt}

Co-design introduces a complex dependency between the design evaluation and policy optimization. %
Formally, our aim is to jointly optimize a morphological design $\omega \in \Omega$ over the design space $\Omega$ and a control policy $\pi_\omega(a | s)$ for this design to perform optimally as measured by the performance over a distribution of target tasks $\mathcal{T}$.
We can define this as the optimization of
\begin{equation}
    \omega^*, \pi^*_{\omega^*} = \argmax_{\omega, \pi_{\omega}} \mathbb{E}_{\pi_{\omega}, \eta \sim \mathcal{T}} \mathcal{L}_{\eta}(\pi_\omega),
\end{equation}
where $\mathcal{L}_\eta$ is a task specific scoring function.

While this can be solved within a single optimization problem, the optimization of policies for redundant mobile manipulators is difficult and prone to local optima, as is defining unbiased evaluators that capture general task performance. In this work, we claim that, for expensive platforms such as mobile manipulators, the cost invested in the optimization and evaluation of direct task performance of strong policies is a valuable tradeoff over the use of cheaper but biased heuristics.

As such, we propose to solve this as a hierarchical decision-making problem, in which we optimize over the design in an outer loop and train a design-specific reinforcement learning policy in the inner loop. This results in an iterative algorithm that solves an MDP for a single sampled design to generate a robust performance score for the design optimization, which we term \ours{}. An overview is shown in \figref{fig:overview}.

\subsection{Policy Optimization}
Given a design $\omega$, our aim is to optimize a policy and produce an indicative estimate of its performance over a wide range of mobile manipulation tasks $\eta \in \mathcal{T}$.
The policy is acting in a design-specific MDP $\mathcal{M}_{\omega, \eta} = \mdp(\mathcal{S}_{\omega, \eta}, \mathcal{A}_\omega, \mathcal{P}_{\omega, \eta}, \mathcal{R}_\eta, \gamma)$ where $\mathcal{S}$ is the state space, $\mathcal{A}$ is the action space, $\mathcal{P}$ is the transition function, $\mathcal{R}_\eta$ is the reward function, and $\gamma \in [0, 1]$ is the discount factor. As we are investigating arm mounting decisions of modular mobile manipulators acting in the same physical world, all tasks share the same underlying state space $\mathcal{S}$ and the same physical dynamics $\mathcal{P}$. But different tasks may act on different objects or motions, resulting in each task acting in a different subspace of $\mathcal{S}$ (in turn implying different dynamics as the state space differs)~\cite{sodhani2021multi}. The design parameters furthermore impact the state space and transition functions through the robot kinematics\myworries{, and potentially the action space}. The objective of the reinforcement learning agent is to find the policy $\pi_\omega(a_t | s_t)$ that maximizes the expected discounted returns over the distribution of tasks
\begin{equation}
    \mathbb{E}_{\pi_{\omega}, \eta \sim \mathcal{T}} \left[ \sum_t{\gamma^t r_{\eta, t}(s_t, a_t)}\right].
\end{equation}

We propose to use a recent state-of-the-art multitask reinforcement learning agent, N$^2$M$^2$~\cite{honerkamp2021learning, honerkamp2023n} that has been demonstrated to outperform optimization-based methods and to perform well across a wide range of tasks in the real world~\cite{honerkamp2024language, schmalstieg2023learning, honerkamp2024zero}. This enables us to train a single agent to perform across the set of tasks, removing the need for an additional optimization loop over tasks.

The N$^2$M$^2$ agent~\cite{honerkamp2023n}, shown in \figref{fig:n2m2}, receives a desired end-effector motion in the base frame of the robot. This motion consists of translational and rotational velocities $\vec{v}_{ee}$ to the next desired pose as well as a more distant end-effector subgoal $g$ in the form of a 6-DoF pose to indicate the longer-term plan. 
The agent then converts these end-effector motions to whole-body motions by generating velocities $\vec{v}_b$ and $v_{torso}$ for the base and torso of the robot. Finally, an inverse kinematics solver completes the motions for the remaining arm joints. The agent is also given control over the speed at which the end-effector motions are executed through a learned scaling factor $||\vec{v}_{ee}||$. \myworries{The agent further observes the current joint positions, its previous actions and a} local occupancy map to learn how to avoid obstacles. At test time, the end-effector plan can be replaced with any arbitrary system to produce motions for new tasks.
We train this agent on a general random goal-reaching task in a procedurally generated obstacle environment that was shown to produce strong generalization~\cite{honerkamp2023n}. \myworries{This environment analytically integrates the velocities of the robot over time and checks for collisions of the robot base and end-effector, enabling very fast simulation without an expensive physics simulator. We then} return the optimized policy $\pi^*_\omega$ for the sampled design to the outer loop. 

\setlength{\tabcolsep}{1pt}
\begin{figure}[t]
	\centering
 \includegraphics[width=.85\linewidth,trim={0cm 0cm 0cm 0cm},clip,angle =0,valign=c]{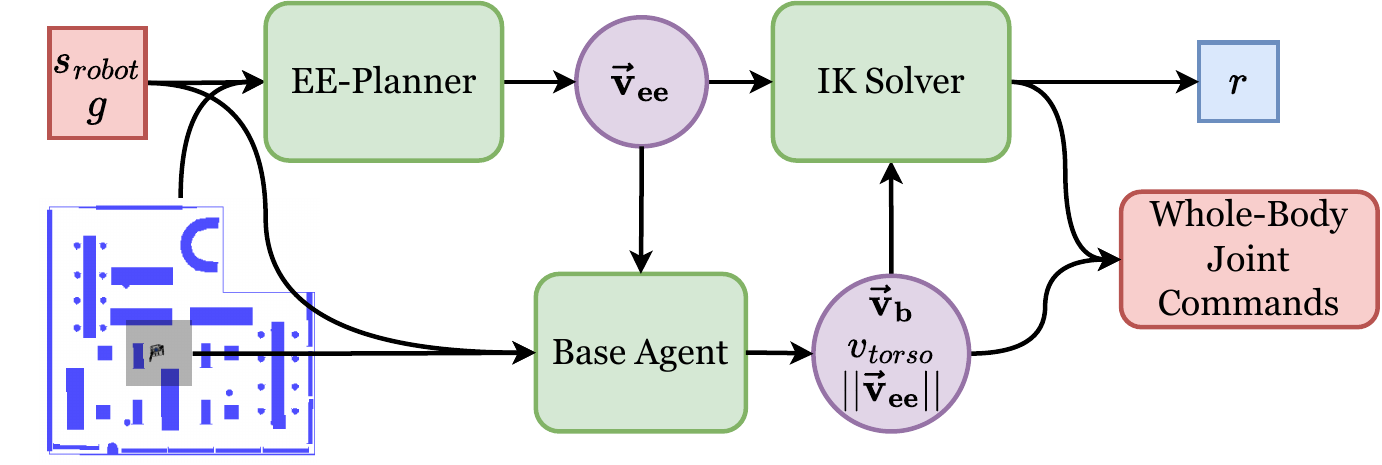}
     \caption{Overview of the N$^2$M$^2$ mobile manipulation policy~\cite{honerkamp2023n}. Given an end-effector motion\myworries{the robot state and a local occupancy map}, a reinforcement learning agent generates controls for the base, torso and the speed of the end-effector motions with the aim to ensure their kinematic feasibility while avoiding any obstacles.\looseness=-1} 
  	\label{fig:n2m2}
\vspace{-0.3cm}
\end{figure}
\setlength{\tabcolsep}{6pt}

\subsection{Task-Driven Design Generation}\label{subsec:design-generation}
In the outer loop, we maintain a distribution over the unknown scoring function $\mathcal{L}: \Omega \xrightarrow{} \mathbb{R}$ that describes the optimality of a given design $\omega$.
We model the distribution $p(\mathcal{L} | D)$ for observed data points $D = \{\omega_i, \mathcal{L}_i \}$ with the Bayesian Optimization and HyperBand (BOHB) algorithm. BOHB combines the advantages of Bayesian optimization's model-based search with HyperBand's efficient resource allocation strategy, making it particularly effective for hyperparameter tuning in high-dimensional design spaces with constrained computational budgets~\cite{falkner2018bohb} while achieving strong final performance. Instead of directly modeling $p(\mathcal{L} | D)$, BOHB maintains a parzen kernel density estimator over $l(\omega) = p(\mathcal{L} < \alpha | \omega)$ and $g(\omega) = p(\mathcal{L} > \alpha | \omega)$ to calculate the expected improvement. Hy\myworries{p}erband then decides on the resource allocation per design by selecting configurations for Successive Halving~\cite{jamieson2016non}. It thereby evaluates many designs with a small budget and increases the budget for the best designs~\cite{falkner2018bohb}.

The BOHB algorithm iteratively updates its distribution \myworries{and} resamples the agent's design parameters to optimize the scoring function. Once new design parameters are generated, they are written to the robot's definition files. The reinforcement learning agent is then relaunched with the updated robot design, entering a new cycle of evaluation and optimization. This continuous feedback loop ensures progressive enhancement of the agent's capabilities in performing the mobile manipulation tasks. We propose to directly use task performance as measured by success rates over a set of representative tasks as a scoring function to measure the general performance of design and policy. Given a set of representative mobile manipulation tasks $\mathcal{T}$, we split it into training tasks $\mathcal{T}_{train}$, validation tasks $\mathcal{T}_{val}$ and held-out test tasks $\mathcal{T}_{test}$. The validation tasks are used to approximate the generalization performance during design optimization, while the test tasks are only used for the evaluation of the final design and policy. This results in a scoring function
\begin{equation}
\mathcal{L}_{val} = \frac{1}{N} \sum_{\eta \in \mathcal{T}_{val}} \nu_\eta,
\end{equation}
where $\nu_\eta$ is the average success rate on task $\eta$. We follow N$^2$M$^2$ and define the success rate as completing the full end-effector motion without violating a maximal translational or orientational tracking error and while avoiding any collisions with the environments.

With this, the independence of the inner and outer optimization loop enables us to decouple the scoring function $\mathcal{L}$ from the dense reward function $\mathcal{R}$ used for RL training. While the dense reward function for RL training is designed to provide an optimal learning signal, it may not be perfectly correlated with the sparse, final task performance.

\subsection{Manipulability Heuristic}\label{sub:manipulability_heuristic}

While our proposed direct task performance metric is highly correlated with the desired outcome, the training of a reinforcement learning agent for each configuration remains costly.
Performance indices have been widely used as simple to compute metrics to summarize the quality of manipulator designs~\cite{xiao2016manipulator, molaei2022kinematic, hwang2017design, feng2020multi}. While less task-specific, this allows the evaluation of a larger set of designs at the same time. As we focus on arm mounting parameters and do not optimize actuators or link lengths, we create an alternative method that uses whole-body manipulability as a scoring function. Yoshikawa's manipulability ellipsoid represents how easily the end-effector can move in any direction, with the volume of the ellipsoid serving as a scalar measure of manipulability~\cite{yoshikawa_manipulability}. It is defined as
\begin{equation}
    m = \sqrt{\det(JJ^T)},
\end{equation}
where $J$ is the Jacobian Matrix and $J^T$ its transpose. We can then define global manipulability over the workspace $W$ as
\begin{equation}
    \mu = \frac{\int m \,dW}{\int dW}.
\end{equation}
We approximate this global manipulability by calculating the manipulability across end-effector poses on a discretized 3D grid centered around the robot and use it as a scoring function in the same optimization loop. \myworries{We extend the manipulability commonly used for manipulator arms to include the based links of the robot.} We base the grid size on the identified task heights in \secref{sec:experiments} and the forward reach of an average human of roughly \SI{0.7}{\meter}~\cite{looker2015reaching}. Including the size of the base, this results in a grid with $x, y, z$ corner coordinates (-0.2, -0.8, 0.1) and (0.2, 0.8, 1.7) relative to the base link of the robot and six orientations per point. For each point, we find an inverse kinematics solution starting from a random configuration and then calculate the manipulability for this configuration. If no inverse kinematics solution exists, we set the manipulability to zero.\looseness=-1 %

To evaluate the efficacy of this heuristic-based optimization, we subsequently train an RL agent using the N$^2$M$^2$ algorithm for the top three designs with the highest global manipulability. This allows us to directly compare the performance of the manipulability-optimized designs with those obtained through the task-based optimization method.

\section{Design Parameters}\label{sec:optimization-parameters}

\begin{figure}%
    \centering
	\resizebox{.9\linewidth}{!}{%
        \includegraphics[trim={0cm 2cm 0cm 0cm},clip,angle =0,valign=c]{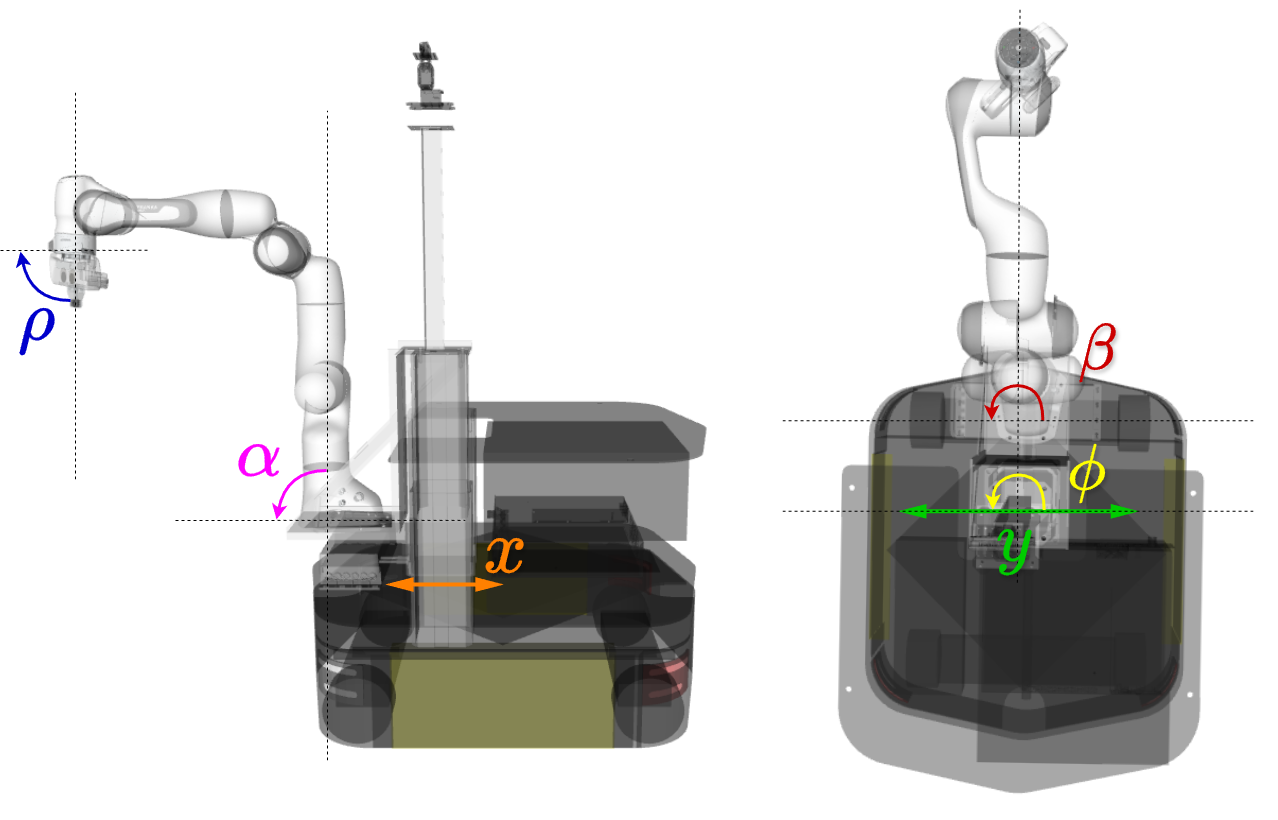}
        }
    \caption{Arm mounting parameters of the design space $\Omega$, cf. \secref{sec:optimization-parameters}, with the rotations $\alpha$ and $\beta$ applied in sequence.}
    \label{fig:optimization_parameters}
    \vspace{-0.3cm}
\end{figure}

We identify the mounting parameters of the existing manipulator arms on mobile robot bases as an important, but often overlooked design aspect of common modular mobile manipulator robots. These parameters have a core impact on the overall kinematic structure. Importantly, they are applicable to a wide range of robots without requiring any change of the individual components, maintaining the cost-efficiency of using existing commercial arms and bases. We identify a set of main mounting parameters based on expert knowledge. These parameters are shown in \figref{fig:optimization_parameters} and consist of the following:

\texttt{Arm Pitch $\alpha$}: forward tilt of the arm base link, ranging from upright (\SI{0}{\degree}) to straight forward (\SI{90}{\degree}).

\texttt{Arm Yaw $\beta$}: yaw angle of the arm base link, ranging from \SI{-90}{\degree} to \SI{90}{\degree}. Applied after rotation $\alpha$, thereby allowing, e.g., mounting of the arm on the side of the tower.

\texttt{End-Effector Pitch $\rho$}: mounting angle of the end effector at the end of the manipulator's arm, ranging from forward (\SI{0}{\degree}) to downward facing (\SI{90}{\degree}).

\texttt{Forward translation $x$}: mounting point of the tower lift link along the x-axis of the robot. We simultaneously shift the position of the arm base link by the opposite value, keeping its absolute position on the robot base the same. This impacts self-collisions between the arm and tower by changing the spacing between them. Limited to \SI{-5}{\cm} to \SI{15}{\cm}.

\texttt{Lateral translation $y$}: translation of the tower lift link along the y-axis of the robot. Constrained to the robot base width of \SI{-20}{\cm} to \SI{20}{\cm}.

\texttt{Tower Yaw $\phi$}: yaw angle of the tower lift link, ranging from  \SI{-90}{\degree} to \SI{90}{\degree}.

This results in a six-dimensional optimization space $\Omega \in \mathbb{R}^6$.
Note that while in this work we focus on arm mounting parameters, the approach is agnostic to the nature of these parameters and applicable to a much wider range of parameters, such as joint structure or link lengths.

\subsection*{Physical Feasibility}\label{sec:tipover}
\begin{table}
    \centering
    \caption{Tipover parameters}
    \label{tab:tipover}
    \begin{threeparttable}
    \begin{tabularx}{\linewidth}{l|rr}
        \toprule
        \textbf{Parameter} & Weight $m$ & Position Vector $\vec{p}$ \\
        \midrule
        \myworries{Franka link 0} & \myworries{\SI{2.40}{\kg}} & \myworries{\mmvec{430}{330}{907}}\\
        \myworries{Franka link 1} & \myworries{\SI{2.79}{\kg}} & \myworries{\mmvec{489}{389}{907}}\\
        \myworries{Franka link 2} & \myworries{\SI{2.54}{\kg}} & \myworries{\mmvec{681}{581}{879}}\\
        \myworries{Franka link 3} & \myworries{\SI{2.25}{\kg}} & \myworries{\mmvec{821}{721}{907}}\\
        \myworries{Franka link 4} & \myworries{\SI{2.20}{\kg}} & \myworries{\mmvec{837}{737}{884}}\\
        \myworries{Franka link 5} & \myworries{\SI{2.29}{\kg}} & \myworries{\mmvec{1005}{965}{880}}\\
        \myworries{Franka link 6} & \myworries{\SI{1.35}{\kg}} & \myworries{\mmvec{1105}{1005}{880}}\\
        \myworries{Franka link 7} & \myworries{\SI{0.36}{\kg}} & \myworries{\mmvec{1113}{1013}{959}}\\
        \myworries{End-effector} & \myworries{\SI{0.71}{\kg}} & \myworries{\mmvec{1260}{1160}{928}}\\
        Max. arm payload & \SI{3.00}{\kg} & \myworries{\mmvec{1260}{1160}{928}}\\ %
        Ewelllix TLT Tower & \SI{22.50}{\kg} & \myworries{\mmvec{300}{200}{280}}\\ %
        Ridgeback Base & \SI{135.00}{\kg} & \mmvec{0}{0}{140} \\
        \bottomrule
    \end{tabularx}
     \begin{tablenotes}[para,flushleft]
       \footnotesize      
       The position vector is the x,y,z coordinate of the COM of the individual component relative to the center of the base, for the configuration with the furthest possible distance from the center (cf. \secref{sec:tipover}). %
     \end{tablenotes}
     \end{threeparttable}
     \vspace{-0.4cm}
\end{table}

As we change the weight distribution of the platforms, we have to ensure the physical feasibility of the resulting designs. The main risk is tipping over if the arm is mounted with a much larger forward reach. We calculate the risk of tipping over both statically and dynamically. We calculate this for the design with the most extreme weight distribution, where the arm is mounted at the edge of the base, with \myworries{tower at maximum height}, the base rotated \SI{45}{\degree} and fully stretched out forward\myworries{, and carrying the maximum possible payload}. The parameters of each component are shown in \tabref{tab:tipover}.

Statically, we approximate the stability of the arm configuration by assuming a point mass at the center of each component. We can then calculate the center of mass as follows:
\begin{flalign}
\text{COM} &= \frac{\sum m_i \times \vec{p}_i}{\sum m_i} = \myworries{(\SI{132}{\mm}, \SI{109}{mm})}.
\end{flalign}
With the four wheels located at $(\pm \SI{319}{mm}, \pm \SI{276}{mm})$, the center of mass lies clearly within the area spanned by the wheels, remaining statically stable. Dynamically, we approximate the case when the robot drives at full speed and \myworries{brakes} abruptly. The critical torque is
\begin{flalign}
    \tau _{critical}=m_{base} \times g \cdot d = \SI{231.76}{\newton\meter},
\end{flalign}
where $m_{base}$ is the mass of the base, $g$ is the weight force and $d = \SI{0.319}{\meter} - \SI{0.144}{\meter}$ is the distance between the rotational axis located at the front wheels of the robot and the center of mass of the robot, as calculated above\footnote{Assuming the worst case design described above would place the rotational axis even further away diagonally at only one of the front wheels. To remain conservative, we continue with the lower estimate of the critical torque.}.
The maximum torque exhibited by the system, $\tau_{max}=\tau_{acc} + \tau_{grav}$, consists of the acceleration torque and the gravitational torque:
\myworries{
\begin{flalign}
    \tau_{acc}&=\sum_{l}m_{l}  \cdot a \cdot z_l \cdot cos (\beta_l) = \SI{21.6}{\newton\meter},\\ %
    \tau_{grav}&=\sum_{l}m_{l}  \cdot g \cdot r_l \cdot sin(\alpha_l) = \SI{136.8}{\newton\meter}, %
\end{flalign}
}
where \myworries{$l$ are the different components given by the arm links and the payload.  %
Respectively, $m_{l}$ are their masses, $r_l$ the distances from the pivot point to their centers of mass, $z_l$ their heights, $\alpha_l$ the angle from the pivot point to the center of each component, $\beta = \SI{90}{\degree} - \alpha$,} and $a=\frac{\Delta v}{\Delta t} = \frac{\SI{1.1}{\meter\per\s}}{\SI{0,5}{\s}}= \SI{2.2}{\meter\per\s\squared}$ is the negative acceleration during the \myworries{braking} process. %
We find that the maximum torque applied is therefore lower than the critical torque, $\tau_{max} < \tau_{critical}$, meaning the system remains stable even under maximal \myworries{braking}.
\myworries{Additionally, the arm pulling on a fixed object can exert an additional torque. We find that the system remains stable even in the case of exerting torques up to the safety limits of \SI{30}{\newton\meter} of the arm. However, this must be carefully considered for arms with higher safety limits.}
\section{Experiments}\label{sec:experiments}
We aim to empirically answer the following questions:
\begin{itemize}
    \item What is the impact of the identified arm mounting parameters for mobile manipulation tasks?
    \item How does the use of direct task metrics compare to the optimization of cheaper-to-evaluate but less correlated performance indices?
    \item What is the impact of the task-distribution $\mathcal{T}_{val}$ used during optimization?
\end{itemize}

\subsection{Tasks}
We parameterize a set of representative mobile manipulation tasks to match common household settings.
The tasks are designed to collectively test the robot’s navigation, object manipulation, and interaction capabilities in scenarios that mimic real-world household activities. Additional details in~\cite{honerkamp2023n}.

\para{\texttt{Random Goal:}} The robot starts at a random position and has to follow a linear end-effector motion from its initial end-effector pose to a random goal pose on an empty map. We draw goals in the range of the human vertical accessibility~\cite{hrovatin2015ergonomic} at the height of \SI{0.1}{\meter} to \SI{1.7}{\meter} (defined for the wrist frame of the robot, resulting in gripper tip poses reaching down to the floor). At the edges of the workspace, we restrict the goal orientations to a pitch matching a top or bottom grasp, respectively. \\ %
\para{\texttt{Random Obstacle (RO):}}
The robot is placed in a procedurally generated obstacle map, requiring navigation in narrow, cluttered environments while following end-effector motions interpolated from an A$^*$-planner towards similar random goals. \myworries{This task was developed to generate highly diverse scenarios, and to learn general behaviors.}\\
\para{\texttt{Pick and Place:}}
The robot has to pick up an object from one table and place it on another table. The pick and place heights are set to the same values as for the random goals, with orientation constraints for top, and bottom grasps at the edges of the workspace height.\\
\para{\texttt{Open Door:}}
The robot has to follow end-effector motions inferred from human demonstrations~\cite{twelsche2017learning} to navigate to a door, press the handle down, and open it while passing through the narrow door frame.\\
\para{\texttt{Open Drawer:}} \myworries{The robot has to} open a drawer with a prismatic joint placed at typical kitchen heights of \SI{0.4}{\meter} to \SI{1.2}{\meter}~\cite{hrovatin2015ergonomic}, following motions from human demonstrations.\\
\para{\texttt{Open Cabinet:}}
Open a cabinet door with a revolute joint, located at typical kitchen heights of \SI{0.4}{\meter} to \SI{1.7}{\meter}~\cite{hrovatin2015ergonomic}, following motions from human demonstrations.

For the \texttt{Pick and Place}, \texttt{Open Door}, \texttt{Open Drawer} and \texttt{Open Cabinet} tasks we also randomly place an obstacle in the path between the robot and the task object, requiring the robot to complete the motions from different directions. \myworries{The motions from human demonstrations adapt to different object poses through their parametrization in an object-centric frame~\cite{twelsche2017learning}.} The tasks are shown in the video.

\subsection{Robots and Optimization Setup}
To test the effectiveness across mobile manipulator platforms, we evaluate all approaches for two drive types, omnidirectional and differential, and two arms, the 7 Degree of Freedom (DoF) Franka and the 6 DoF UR5 arm. 
To evaluate the impact of these different modules, we substitute them into the same base design of the FMM robot, shown in \figref{fig:teaser}. The default configuration of this robot consists of an omnidirectional base, a height-adjustable Ewellix column \myworries{to change the height}, and a Franka arm.\looseness=-1

All hyperparameters of the optimization process are reported in \tabref{tab:bohb_parameters}.
We set the budget for Hy\myworries{p}erband between 300,000 and 1M training steps. These values are chosen based on the average success rate of the original N$^2$M$^2$ agent~\cite{honerkamp2023n}. Approximately after 200,000 training steps, the curve starts to rise, converging at around 700,000 steps. We set the boundaries optimistically to make sure we find the best performing agents. The parameters of the N$^2$M$^2$ agent are set according to~\cite{honerkamp2023n}, which were shown to generalize across robot platforms.

\begin{table}
    \centering
    \caption{BOHB Algorithm Parameters. }
    \label{tab:bohb_parameters}
    \begin{threeparttable}
    \begin{tabularx}{.65\linewidth}{l|c}
        \toprule
        \textbf{Parameter} & \textbf{Value} \\
        \midrule
        Iterations & 20 \\
        Sampled Designs & 60 \\
        Reduction Factor $\eta$ & 3 \\
        Random Fraction $\rho$ & 1/3 \\
        Minimum Budget $b_{min}$ (steps) & 300,000 \\
        Maximum Budget $b_{max}$ (steps)& 1,000,000 \\
        \bottomrule
    \end{tabularx}
     \begin{tablenotes}[para,flushleft]
       \footnotesize      
       All other parameters are set to the default values of the HpBandSter framework \cite{falkner2018bohb}.
     \end{tablenotes}
     \end{threeparttable}
    \vspace{-0.4cm}
\end{table}

\subsection{Models}
\para{Tabletop Mount}: The default tabletop-based configuration of the arm without any design optimization.\\
\para{Manipulability}: Co-design based on the global manipulability as described in \secref{sub:manipulability_heuristic}.\\
\para{\ours{} (RO Task)}: Our approach with $\mathcal{T}_{train} = \mathcal{T}_{val} = \{\texttt{Random Obstacle}\}$. To ensure the designs learn to pass narrow passages, we reduce the spacing between obstacles during training to \SI{1.7}{\meter}.\\
\para{\ours{} (All Tasks)}: Our approach with $\mathcal{T}_{train} = \{\texttt{Random Obstacle}\}$ and $\mathcal{T}_{val} = \mathcal{T}_{test} = \{\texttt{All tasks}\}$\myworries{, equally weighted}.

We run all methods with the same time budget of roughly eight days on a server with a Nvidia A40 GPU and AMD EPYC 7513 processor, resulting in the evaluation of 60 designs for \ours{} and 2,500 designs for the cheaper manipulability approach. Note that these compute costs remain negligible compared to the hardware cost of the platforms.

\subsection{Experimental Results}

\begin{sisetup}{round-mode=places, round-precision=2}
\begin{table*}
    \centering
    \caption{Evaluation of the generated designs across tasks with standard error.}
    \label{tab:evaluation_res_conclusion}
    \begin{threeparttable}
    \begin{tabularx}{\linewidth}{p{0.3cm}p{0.3cm}l|YYYYYY|Y||Y}       
         \toprule
         \centering Drive & \centering Arm & Model & Random Obstacle & Random Goal & Pick~\& Place & Door & Drawer & Cabinet & Average Success & Manipu-lability \\ 
         \midrule
         \parbox[t]{3mm}{\multirow{4}{*}{\rotatebox[origin=c]{70}{Omni}}} & \parbox[t]{3mm}{\multirow{4}{*}{\rotatebox[origin=c]{70}{Franka}}} 
         & Tabletop Mount & 30 \myworries{$\pm$ 4.6} & 54 \myworries{$\pm$ 5.0} & 39 \myworries{$\pm$ 4.9} & \phantom{0}6 \myworries{$\pm$ 2.4} & 36 \myworries{$\pm$ 4.8} & 26 \myworries{$\pm$ 4.4} & 31.8 \myworries{$\pm 1.9$} & \num{1.154616224434318}\\
         && Manipulability & 54 \myworries{$\pm$ 5.0} & 73 \myworries{$\pm$ 4.4} & 33 \myworries{$\pm$ 4.7} & 18 \myworries{$\pm$ 3.8} & 54 \myworries{$\pm$ 5.0} & 53 \myworries{$\pm$ 5.0} & 47.5 \myworries{$\pm 2.0$} & \num{1.448003469407171}\\
         && \ours{} (RO task) & \textbf{66} \myworries{$\pm$ 4.7} & \textbf{91} \myworries{$\pm$ 2.9} & \textbf{51} \myworries{$\pm$ 5.0} & 57 \myworries{$\pm$ 5.0} & \textbf{83} \myworries{$\pm$ 3.8} & \textbf{88} \myworries{$\pm$ 3.2} & \textbf{72.7} \myworries{$\pm 1.8$} & \num{1.1814370816792519} \\
         && \ours{} (all tasks) & 44 \myworries{$\pm$ 5.0} & 74 \myworries{$\pm$ 4.4} & 30 \myworries{$\pm$ 4.6} & \textbf{66} \myworries{$\pm$ 4.7} & 74 \myworries{$\pm$ 4.4} & 66 \myworries{$\pm$ 4.7}  & 59.0 \myworries{$\pm 2.0$}& \num{0.8832559906244507} \\
         \midrule
         \parbox[t]{3mm}{\multirow{4}{*}{\rotatebox[origin=c]{70}{Differential}}} & \parbox[t]{3mm}{\multirow{4}{*}{\rotatebox[origin=c]{70}{Franka}}} 
         & Tabletop Mount & 16 \myworries{$\pm$ 3.7} & 38 \myworries{$\pm$ 4.9} & \phantom{0}6 \myworries{$\pm$ 2.4} & \phantom{0}1 \myworries{$\pm$ 1.0} & 15 \myworries{$\pm$ 3.6} & \phantom{0}5 \myworries{$\pm$ 2.2} & 13.5 \myworries{$\pm 1.4$}& \num{0.63912506178942164}\\
         && Manipulability & 41 \myworries{$\pm$ 4.9} & 84 \myworries{$\pm$ 3.7} & 36 \myworries{$\pm$ 4.8} & 24 \myworries{$\pm$ 4.3} & 47 \myworries{$\pm$ 5.0} & 10 \myworries{$\pm$ 3.0} & 40.3 \myworries{$\pm 2.0$} & \num{1.4517653822701372}\\
         && \ours{} (RO task) & \textbf{57} \myworries{$\pm$ 5.0} & \textbf{92} \myworries{$\pm$ 2.7} & \textbf{37} \myworries{$\pm$ 4.8} & 31 \myworries{$\pm$ 4.6} & \textbf{59} \myworries{$\pm$ 4.9} & \textbf{49} \myworries{$\pm$ 5.0} & \textbf{54.2} \myworries{$\pm 2.0$}& \num{1.3251901743996877}\\
         && \ours{} (all tasks) & 23 \myworries{$\pm$ 4.2} & 75 \myworries{$\pm$ 4.3} & 23 \myworries{$\pm$ 4.2} & \textbf{55} \myworries{$\pm$ 5.0} & 25 \myworries{$\pm$ 4.3} & 25 \myworries{$\pm$ 4.3}  & 37.6 \myworries{$\pm 2.0$}& \num{1.1864994527768962}\\
         \midrule
         \parbox[t]{3mm}{\multirow{4}{*}{\rotatebox[origin=c]{70}{Omni}}} & \parbox[t]{3mm}{\multirow{4}{*}{\rotatebox[origin=c]{70}{UR5}}} 
         & Tabletop Mount & 26 \myworries{$\pm$ 4.4} & 46 \myworries{$\pm$ 5.0} & \phantom{0}9 \myworries{$\pm$ 2.9} & \phantom{0}4 \myworries{$\pm$ 2.0} & 26 \myworries{$\pm$ 4.4} & \textbf{34} \myworries{$\pm$ 4.7} & 24.2 \myworries{$\pm 1.7$}& \num{0.050350043818548275}\\
         && Manipulability & 28 \myworries{$\pm$ 4.5} & \textbf{57} \myworries{$\pm$ 5.0} & \phantom{0}9 \myworries{$\pm$ 2.9} & 15 \myworries{$\pm$ 3.6} & \textbf{32} \myworries{$\pm$ 4.7} & 33 \myworries{$\pm$ 4.7} & 29.0 \myworries{$\pm 1.9$}& \num{0.52537005520}\\
         && \ours{} (RO task) & \textbf{35} \myworries{$\pm$ 4.8} & 47 \myworries{$\pm$ 5.0} & 17 \myworries{$\pm$ 3.8} & 16 \myworries{$\pm$ 3.7} & 28 \myworries{$\pm$ 4.5} & 17 \myworries{$\pm$ 3.8} & 26.7 \myworries{$\pm 1.8$}& \num{0.18184924965597238}\\
         && \ours{} (all tasks) & 24 \myworries{$\pm$ 4.3} & 44 \myworries{$\pm$ 5.0} & \textbf{22} \myworries{$\pm$ 4.1} & \textbf{19} \myworries{$\pm$ 3.9} & \textbf{32} \myworries{$\pm$ 4.7} & \textbf{34} \myworries{$\pm$ 4.7} & \textbf{29.2} \myworries{$\pm 1.9$}& \num{0.11215272338759553}\\
         \bottomrule
    \end{tabularx}
     \begin{tablenotes}[para,flushleft]
       \footnotesize      
       Left: success rates in \%, best highlighted in bold. Right: average manipulability of the designs as defined in \secref{sub:manipulability_heuristic}. RO task: optimization on the random obstacle task. \myworries{$\pm$ indicates estimated standard errors.}
     \end{tablenotes}
     \end{threeparttable}
\vspace{-0.3cm}
\end{table*}
\end{sisetup}

We evaluate the resulting policies and designs over 100 episodes per test task and report the success rates in \tabref{tab:evaluation_res_conclusion}. All approaches achieve improvements over the original tabletop mounting design, confirming the importance of the arm mounting parameters. \ours{} achieves the best results across platforms, more than doubling success rates for the Franka arm. In contrast, while able to sample many more designs in the same time, the designs generated by the manipulability heuristic do not reach the same level of improvement. We report the manipulability for all final designs on the right of \tabref{tab:evaluation_res_conclusion}. The numbers confirm our hypothesis that the performance index is only weakly correlated with task performance, with a Pearson correlation coefficient of $\rho=0.58$. Moreover, the definition of the workspace over which to evaluate the metric further biases the designs, e.g., a symmetric workspace induces a prior to cover all directions. \myworries{While a control policy could move the base to account for one-sided designs, the manipulability can only take into account the local impact of the base on the Jacobian, not larger repositionings.}

Comparing the evaluation of \ours{} validated on all tasks versus the general random obstacle task, we find that the random obstacle task results in significantly better designs for the two Franka-based platforms, with small advantages of incorporating all tasks for the UR5 arm.
We hypothesize that this effect is partially caused by negative correlations between certain tasks, leading to a more fragmented design optimality function that would require more data to discover global optima. \myworries{It furthermore relates to the generality of the random obstacle task, which was designed for diverse experience, in contrast to the application-driven downstream tasks. Nevertheless, it also} points to further possible improvements in the design sampling approach. Improvements are generally smaller for the UR5 arm. This hints at the configuration space already being better aligned with the mobile manipulation tasks than the Franka arm, which was specifically designed for tabletop tasks. At the same time, overall performance on the tasks also remains clearly below the Franka arm, and so does the average manipulability. This demonstrates the importance of the additional DoF of the Franka arm.
\myworries{We also evaluate the share of steps in which the arm is in a near singular configuration. We find no significant increases, except for the cabinet task of the RO trained agents with the Franka arm. However, the sucess rates remain by far the highest, indicating that the agent is successfully using the mobility of the base to compensate for limitations of the arm.}

\begin{figure}
    \centering
	\resizebox{\linewidth}{!}{%
\includegraphics[width=\linewidth,trim={0cm 2.7cm 0cm 0cm},clip,angle =0,valign=c]{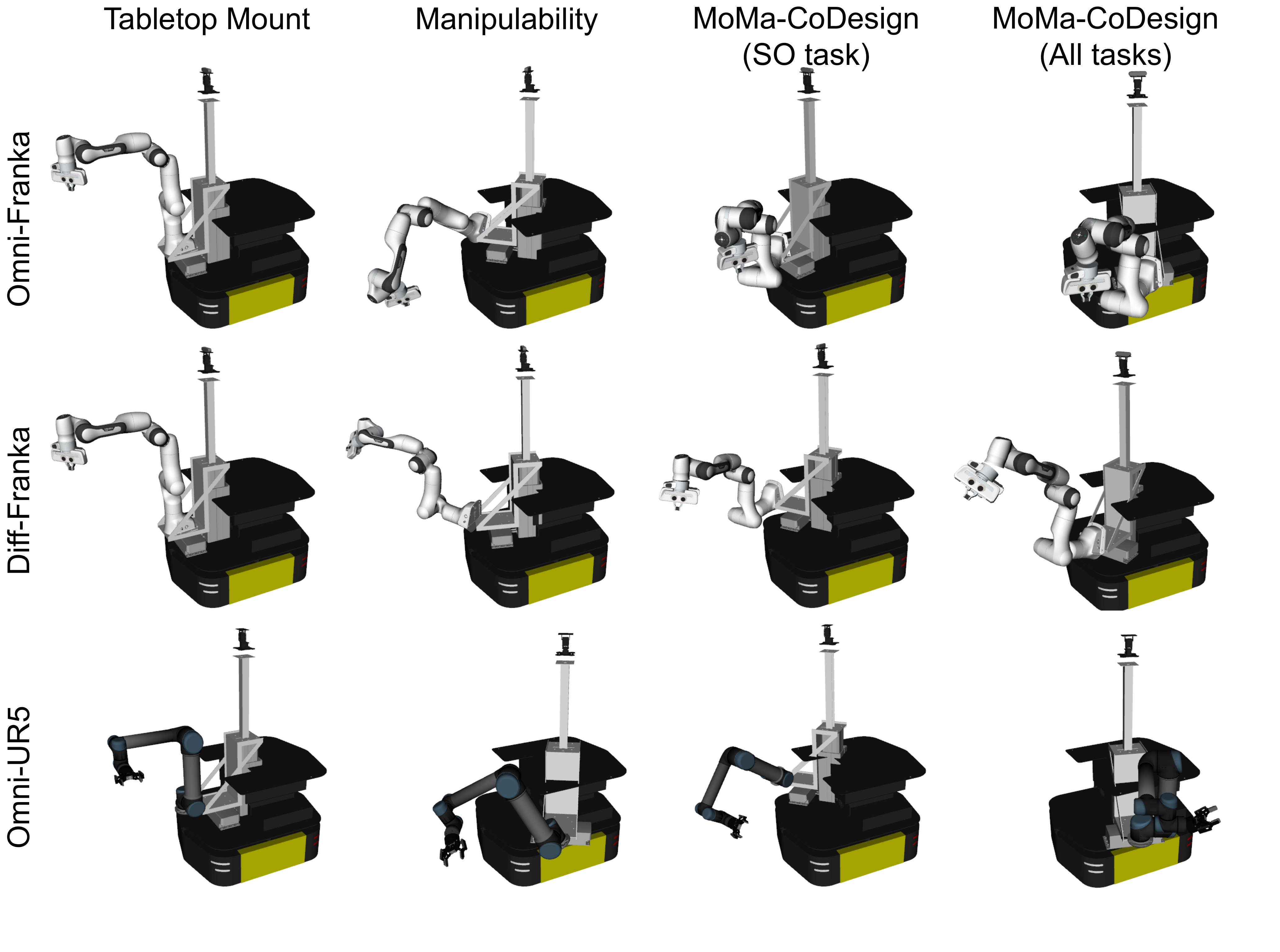}
        }
    \caption{Designs generated by the different methods across the different robots. Note that we do not render connecting links between tower and arm and between arm and end-effector.}
    \label{fig:designs}
    \vspace{-0.3cm}
\end{figure}

The designs generated by the different approaches are shown in \figref{fig:designs}. We find that all designs prefer to tilt the arm forward and place the tower towards a corner of the base, increasing its reach. The resulting designs differ strongly from the tabletop mounting configuration, again highlighting the importance of this design factor. At the same time, the orientations of the arm significantly differ between the manipulability-optimized designs and \ours{}. Furthermore, we find that \ours{} converged to similar designs across the four Franka arm configurations, showing that the design optimization process moved towards a similar optimum.
We show the designs performing the different tasks in the accompanying video. 
\begin{figure}
    \centering
	\resizebox{\linewidth}{!}{%
	\begin{tabular}{cccc}
        \includegraphics[scale=1, trim={0cm 0cm 0cm 0cm},clip,angle =0,valign=c]{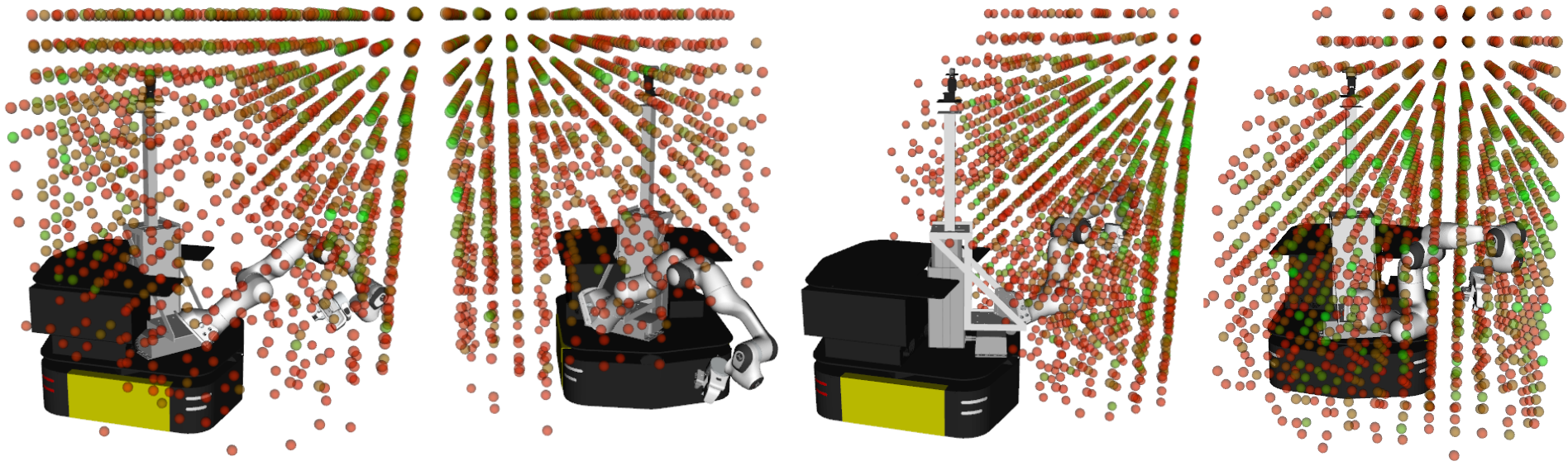}
	    \end{tabular}
        }
    \caption{Manipulability across the workspace of the design for the omnidirectional Franka robot. \myworries{Left: design generated by optimizing the manipulability, right: design generated by \ours{} (SO task). Green} indicates high manipulability, and \myworries{red} is low. Points with zero manipulability are omitted. Each point depicts the average over six orientations.}
    \label{fig:heatmap}
    \vspace{-0.3cm}
\end{figure}
\figref{fig:heatmap} shows the manipulability for the design generated by the manipulability method. As we can see, the resulting design covers large areas of the workspace. However, as we showed previously, this does not directly correlate with the required workspace and motions for diverse mobile manipulation tasks.

\section{Conclusion}
In this work, we identify arm mounting parameters as an important design factor for modular mobile manipulation platforms. We showed that default tabletop-like mounting configurations are suboptimal for a wide range of mobile manipulation tasks that often require different workspaces and motions at different heights. To address these problems, we presented a co-design approach that optimizes a powerful reinforcement learning agent directly on task metrics. We then empirically demonstrated the approach across different base and arm modules. In comparison to performance heuristics, we find that the designs generated by our method lead to significantly higher performance, as heuristics are only weakly correlated with the objective of interest, and as a result, even with the evaluation of a much larger number of designs, converge to worse optima. Finally, we open-sourced the resulting designs and approach to enable the community to test and generate alternative designs for different base and arm modules. \myworries{In future work, we plan to investigate the implementation of the resulting designs in the real world, as well as the use of heuristic-based methods to initialize the search of the presented co-design approach through computationally cheaper methods.}

\footnotesize
\bibliographystyle{IEEEtran}
\bibliography{IEEEabrv,bibliography.bib}

\end{document}